\documentclass{article}

\usepackage{arxiv}

\usepackage[utf8]{inputenc} 
\usepackage[T1]{fontenc}    
\usepackage{hyperref}       
\usepackage{url}            
\usepackage{booktabs}       
\usepackage{amsfonts}       
\usepackage{nicefrac}       
\usepackage{microtype}      
\usepackage{graphicx}
\usepackage{natbib}
\usepackage{doi}
\usepackage{amsmath}
\usepackage{multirow}
\setcitestyle{authoryear, open={(},close={)}}
\newcommand{\beginsupplement}{%
        \setcounter{table}{0}
        \renewcommand{\thetable}{S\arabic{table}}%
        \setcounter{figure}{0}
        \renewcommand{\thefigure}{S\arabic{figure}}%
     }

\title{Depth and Representation in Vision Models}

\author{ \href{https://orcid.org/0000-0003-1661-4579}{\includegraphics[scale=0.06]{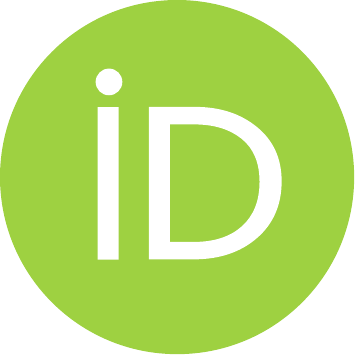}\hspace{1mm}Benjamin L. Badger}\thanks{The author would like to thank Guidehouse for support during the research and writing of this paper.  Code for this work may be found on \url{https://github.com/blbadger/depth-representation}.} \\
	Guidehouse \\
	1200 19th St. NW Washington, DC 20036 \\
	\texttt{bbadger@guidehouse.com} \\
}

\date{}

\renewcommand{\headeright}{Technical Report}
\renewcommand{\undertitle}{Technical Report}

\hypersetup{
pdftitle={Depth and Representation in Vision Models},
pdfsubject={Deep Learning},
pdfauthor={Benjamin L. Badger},
pdfkeywords={Deep Learning, Representation Theory},
}

\begin{document}
\maketitle

\begin{abstract}
	Deep learning models develop successive representations of their input in sequential layers, the last of which maps the final representation to the output. Here we investigate the informational content of these representations by observing the ability of convolutional image classification models to autoencode the model's input using embeddings existing in various layers. We find that the deeper the layer, the less accurate that layer's representation of the input is before training. Inaccurate representation results from non-uniqueness in which various distinct inputs give approximately the same embedding.  Non-unique representation is a consequence of both exact and approximate non-invertibility of transformations present in the forward pass.  Learning to classify natural images leads to an increase in representation clarity for early but not late layers, which instead form abstract images.  Rather than simply selecting for features present in the input necessary for classification, deep layer representations are found to transform the input so that it matches representations of the training data such that arbitrary inputs are mapped to manifolds learned during training. 
	
\end{abstract}


\section{Introduction}
    Deep learning models are termed deep because they are composed of sequential layers which pass input information to each other in succession to form an output. The theory of deep learning postulates that each layer forms a representation of the input such that certain features of the input are selected, and these representations become more abstract as model depth increases \citep{zeiler2014visualizing}. The last layer provides a mapping from these abstract features to the output, and the goal of forming representations of the input in successive layers is to make that mapping process effective as possible. Presumably this is done by passing only the information necessary to allow the final layer's mapping to approximate the desired output.
    
    There is abundant evidence to support this theory of deep learning representations \citep{goodfellow2016deep, oquab2014}, which is in some sense somewhat counter-intuitive given the architectures of commonly used vision models.  Focusing on convolutional neural networks used for image classification, it is unclear that increasing layer depth would necessarily limit the information that is passed from input to output given that many models currently in use contain more elements in each layer than there are elements in the input.  Embedding theory postulates that a model typically requires a reduction in the number of nodes in a hidden layer relative to the input in order to perform a proper (non-trivial) embedding such that the input is not merely copied.
    
    Perhaps the most direct way to understand embedding informational content is to observe the ability of embeddings (which are equivalent to a layer outputs) of various hidden layers of vision models to copy the model's input. In this work the information present in a layer embedding is observed via performing gradient descent on an initially random input in order to give an output from that layer that approximates the output of that same layer given some target image input.  This procedure may be thought of as performing an autoencoding of the input, with the similarity (via a distance metric or simply a visual measure) between the input generated and the true input being a measure of how accurate that autoencoding is.  Assuming that the gradient descent procedure is sufficiently powerful to generate arbitrary images, a better autoencoding signifies more information exists in the layer's embedding with respect to the input whereas a worse autoencoding signifies the opposite.
    
    The question of whether input information is restricted in successive layer representations is addressed first, and then a theory behind the answer to this question is developed, as well as some insight into how the training process informs layer representations.  
    
\subsection{Related Work}
    Elsewhere, shallow histograms of oriented gradient representation visualizations were used to understand instances of mis-identification \citep{vondrick2013hoggles} which drew from earlier work on visualization of image matching features in the context of layered difference-of-Gaussian images \citep{lowe2004distinctive}.  More recently, work which used gradient-based updates on initially random inputs to visualize convolutional neural network hidden layer representations \citep{mahendran2015understanding} focused on invariances between representation visualizations using various regularizers in the gradient update.  It was also shown that deep learning image representations are capable of separating image style from content (or equivalently abstract features from specific ones), allowing for the gradient-based generation of an image that has the content of one target while applying the artistic style of another target image \citep{gatys2016image}.  

\subsection{Our Contribution}
    The point of departure for this work is that rather than attempting to make the most accurate visualization for some deep layer representation, or aiming to manipulate an image using representations, we instead focus on the limits of representations and specifically focus on their informational content.  Investigating this question we observe both trained and untrained vision model representations and the specific changes that occur upon training.  We find that representations become more and more inaccurate as model depth increases, and provide a theoretical basis for how this occurs due to greater non-uniqueness with increased depth. We provide an alternative rationale for why deep vision models tend to learn Gabor filters in their early layers of these models, and find experimental support for the idea that later layers impose their expectations of what the input should be on their representations, rather than simply attempting to select for certain features of that input.  We conclude that vision models trained for classification tend to generate their own interpretations of any given input even if they are not specifically trained to do so, supporting the hypothesis that image recognition and generation are required for one another \citep{hinton2007recognize}.
     
\section{Representation Accuracy Decreases with Depth}
    
    In an attempt to understand the informational content in various layers and whether this information is restricted the deeper the layer in question, we take the direct approach of visualizing the embedding that exists in that layer.  This visualization may be accomplished in a variety of ways, but here we choose a method that is particularly germane to the learning process: gradient descent. 
    
    Given some input $a$, a model designated by parameters $\theta$ and a number of model layers composed as the function $O_l$, the output at the layer of interest is denoted in Equation (\ref{eq1}).
    
    \begin{equation}
    \hat{y} = O_l(a, \theta) 
    \label{eq1}
    \end{equation}
    
    This $\hat{y}$ vector is the representation in layer $l$ of input $a$ and contains the information that layer has with respect to the input.  Visualizing this representation allows for the understanding of the information it contains, so what we want to find is some input $a_n$ that gives an output at the same layer that approximates the target output $\hat{y}$ as shown in (\ref{eq2}).
    
    \begin{equation}
    y = O_l(a_n, \theta)
    \label{eq2}
    \end{equation}
    
    We can use any number of objective functions to compare the similarity of $\hat y$ and the output $y$ for some other input $a_n$, but here we choose the $L^1$ metric.  The gradient of this metric with respect to the input is given in (\ref{eq3}) for elements indexed by $i$ in output $y$.
    
    \begin{equation}
    g = \nabla_{a_n} \sum_i |\hat{y_i} - y_i| 
    \label{eq3}
    \end{equation}
    
    The choice of initial input $a_0$ is somewhat arbitrary, and for this work we assign pixels randomly on a scaled Gaussian distribution such that $a_0 = \mathcal{N}(a; \;\mu= 7/10, \; \sigma=1/20)$.  
    
    The gradient in (\ref{eq3}) may be used to update the input at iteration $n$ directly as shown in Equation (\ref{eq5}), or applied with a $\mathbb{R}^{3x3}$ Gaussian convolution $\mathcal{N}_c$ at each iteration as in (\ref{eq6}), both of which may be implemented with an $\epsilon$ that decreases as $n$ increases or else stays constant.  If Gaussian convolution is applied, the distribution's standard deviation $\sigma$ typically is assigned to start at $\sigma = 2.4$ (pixels) before linearly decreasing to $\sigma = 0.4$ as $n \to f$ where $f$ is the number of iterations specified.
    
    \begin{equation}
    a_{n+1} = a_n - \epsilon * g
    \label{eq5}
    \end{equation} 
    
    After $n=f$ iterations of (\ref{eq5}) or (\ref{eq6}), the generated input is termed $a_g$.
    
    \begin{equation}
    a_{n+1} = \mathcal{N}_c(a_n - \epsilon * g)
    \label{eq6}
    \end{equation}
    
    This visualization technique is a variation on those have been used extensively for output class visualization \citep{Auduno2015}, feature visualization \citep{olah2017feature}, and deep dream \citep{Dream2015}.  
    
    First we will explore  representations of an image of a dalmatian, one of the 1000 classes of the ImageNet 1K dataset (hereafter referred to as 'ImageNet'), a commonly used excerpt from the full ImageNet dataset \citep{Deng2009}.  This image is chosen because it contains multiple levels of recognizable detail: the general outline of the canine as well as the specific pattern of spots on its coat.  We focus on the ResNet class of models \citep{He2016}, reasoning that the residual connections in these models would make gradient descent on the input easier to tune, ie that there would be a wider range of possible $\epsilon$ values that yield a sufficiently small $|\hat{y} - y|$.

\begin{figure}[h]
        \centering
        \includegraphics[width=0.9\textwidth]{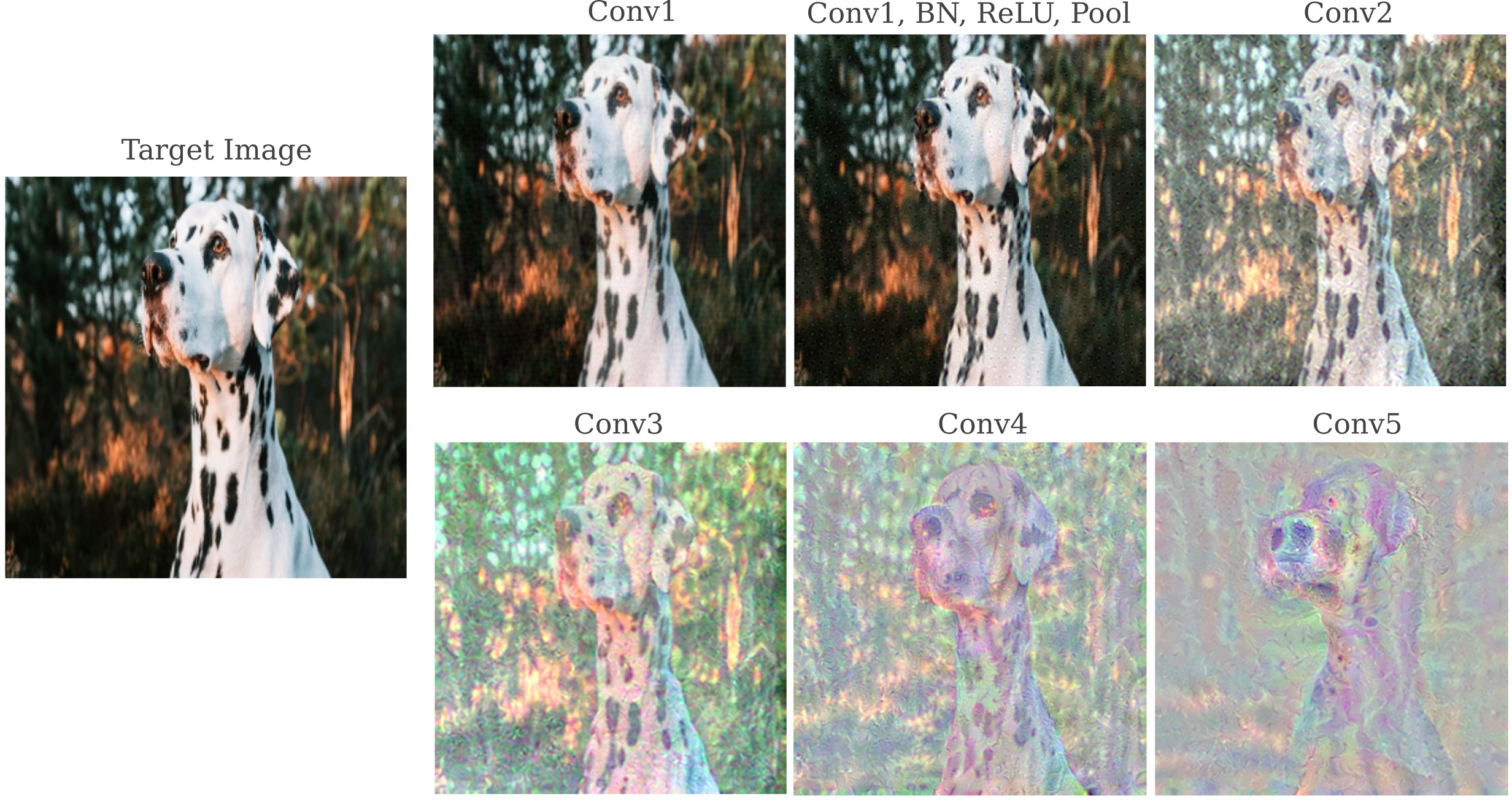}
        \caption{Trained ResNet50 representations of a Dalmatian.  Decreasing $\epsilon$ and constant $n$ used with 3x3 Gaussian convolution per iteration, and the embeddings at the output of each convolutional block as denoted is visualized.}
        \label{fig1}
    \end{figure}

    Applying (\ref{eq6}) to a random Gaussian input, the representation visualizations obtained for certain layers of ResNet50 after a fixed number of iterations $n$ are shown in Figure \ref{fig1}. We follow the naming configuration in the original ResNet publication \citep{He2016} rather than the PyTorch implementation which renames layers Conv2 as Layer 1, Conv3 as Layer 2 etc.
    
    It is clear that the representation becomes more abstract the deeper the layer in question, with a near-perfect copy of the input image in the early layers giving way to a cartoon-like abstraction in the later layers. Of note, the general outline of some canine features (head, nostrils) are observed in all layers but finer details like the exact spot pattern of the input are only found from layers Conv1 to Conv3, whereas the representations from Conv4 and Conv5 contain a noticeably different spot pattern than the original input.
    
    Each representation may be viewed as a combination of an exact copy of the input (a `trivial' representation) and what could be thought of as a true representation, one that is the result of the learning procedure.  To investigate the relative contributions of each type of representation to each layer, we applied (\ref{eq6}) using an untrained ResNet50 model with the reasoning that an untrained model's layer representations would have no contributions from the `true' representation category.  
    
    \begin{figure}[h]
        \centering
        \includegraphics[width=0.65\textwidth]{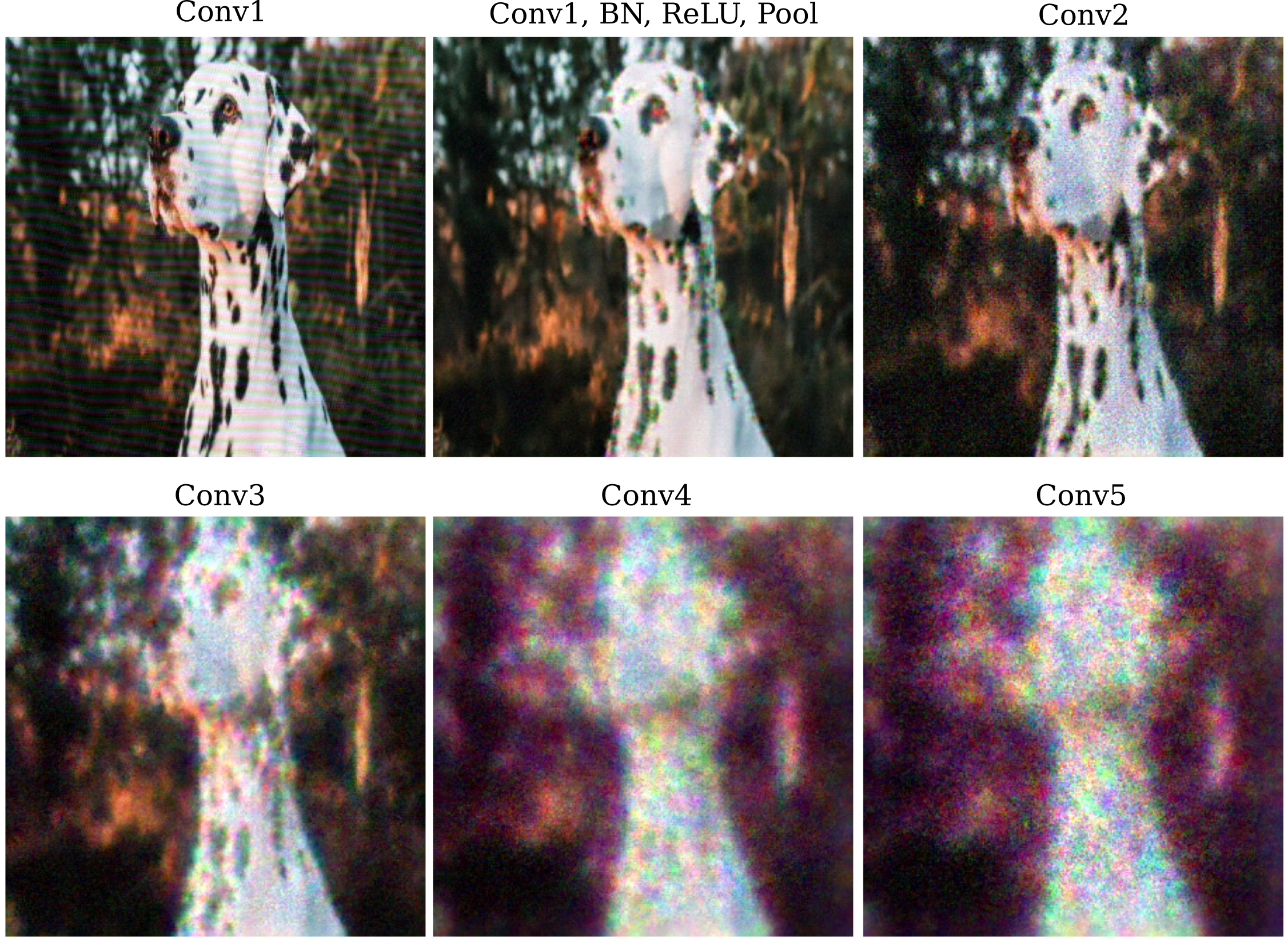}
        \caption{Untrained ResNet50 Input Representation, matching the conditions in Figure \ref{fig1}.}
        \label{fig2}
    \end{figure}
    
    From the results shown in Figure \ref{fig2}, it is clear that the unlearned representations provide less and less information as layer depth increases.  Notably for this particular input, the spot pattern on the Dalmatian's coat disappears in the same layer (Conv4) of the untrained model that the new spot pattern appears in the trained model's representation (Figure \ref{fig1}).  The same observations are made when (\ref{eq5}) is used, meaning that this decrease in input information present in the later layers does not result from Gaussian convolution during the representation visualization process (Figure \ref{figs3}).
    
\section{Imperfect Representations result from Non-Uniqueness}

\subsection{Visually poor input representations despite accurate embedding approximation}

    The inability of untrained image recognition models to form an accurate representations of an input is unexpected given the immense size of these models.  For example, layer Conv4 in ResNet50 has over $369,000$ output elements, whereas the input it fails to accurately approximate (Figure \ref{fig2}) has only $3*299*299=268,203$ elements.  Understanding why these deep learning models are incapable of making accurate trivial representations despite having more than enough parameters to do so is investigated next in depth.
    
    The process of representation visualization, ie the transformations necessary to send $a_0 \to a_g$, may be thought of as many iterations of three separable components: first a forward pass is performed to obtain some output via (\ref{eq1}) followed by comparison to the target output obtained in (\ref{eq2}) to obtain the gradient (\ref{eq3}), and finally this gradient is used to update the input via gradient descent with (\ref{eq5}) or without (\ref{eq6}) Gaussian convolution.  To understand how the representation fails to approximate an input for deeper layers, we consider each component of the visualization process.
    
    First we consider whether or not the gradient acquisition and update components are the cause of the poor representations.  To do this, we test whether the output of $a_g$ approximates the output of $a$ sufficiently.  Rather than use the $L^1$ metric which is minimized during gradient descent, we employ an $L^2$ metric to avoid bias.  Specifically the $L^2$ norm of the difference between the target embedding and generated image embedding, $m_g$, is computed (\ref{eq7}) which is analogous to an n-dimensional version of the Frobenius norm of the difference between output tensors.  This is identical to converting the output tensors into vectors by reshaping them into $\Bbb R^{ix1}$ matricies, where $i$ signifies the number of elements in each output tensor (both of which must be the same shape), and then finding the $L^2$ norm on the difference between these vectors.
    
    \begin{equation}
        m_g = ||O(a_g, \theta) - O(a, \theta)||_2 = \sqrt{\sum_i \left( O(a_g, \theta)_i - O(a, \theta)_i \right)^2}
        \label{eq7}
    \end{equation}
    
    The value of $m$ without some reference is not very illuminating given the abstract nature of hidden layer architectures.  This reference may be supplied by choosing a slightly shifted value $a'$ by adding random normally distributed values to $a$ as shows in (\ref{eq8}).
    
    \begin{equation}
        a' = a + \mathcal{N}(a; \mu=0, \sigma=1/20)
        \label{eq8}
    \end{equation}
    
    One may establish a measure of the distance $m_r$ between the outputs given the shifted input and original as shown Equation (\ref{eq9}), which is compared to $m_g$ with the assumption that $m_g < m_r$ signifies a `good' approximation if $a'$ is sufficiently close to $a$.  
    
    \begin{equation}
        m_s = ||O(a', \theta) - O(a, \theta)||_2
        \label{eq9}
    \end{equation}
    
\begin{figure}[h]
        \centering
        \includegraphics[width=0.9\textwidth]{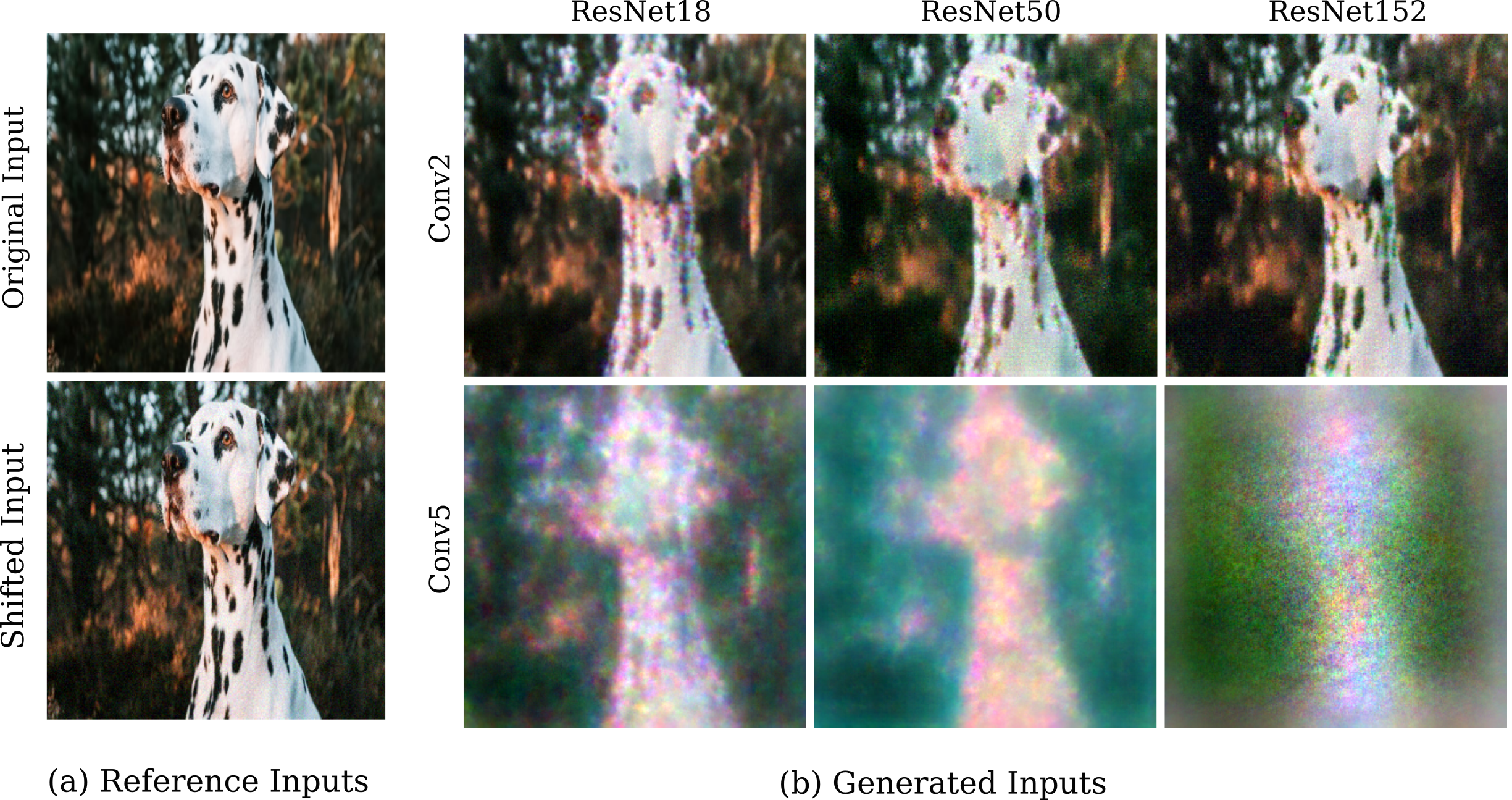}
        \caption{Untrained ResNet50 Input Representations. (a) Reference inputs. (b) Generated inputs $a_g$ with a fixed number of iterations. For all images, $m_g < m_s$.}
        \label{fig3}
    \end{figure}

     For two layers of ResNet50, we see that this is indeed the case: in Figure \ref{fig3} each representation visualization exhibits an $m_g < m_s$ regardless of visual quality, and the same is true for the same layers ResNet152 and for ResNet18.  Observing the difference in Conv5 representation visual quality between ResNet50 and ResNet152, it is clear that an increase in depth further reduces the input representation visualization quality regardless of layer width (Figure \ref{fig3}).
    
\subsection{Analytically poor input representations}
    
    Thus it appears that the gradient descent procedure is capable of at least one kind of accurate approximation.  It may be wondered whether this approximation to within a value of $m$ of $O(a, \theta)$ using an $L^2$ metric is sufficient to say that our gradient descent procedure is actually effective. Instead of comparing $m_g = ||O(a_g, \theta) - O(a, \theta)||_2$ to some reference value, we can instead we can try to understand if our approximation of $\hat y = O(a, \theta)$ is sufficient by comparing the limit of the output (embedding) distance represented by (\ref{eq9}) to the limit of the input distance represented by Equation (\ref{eq10}) as the gradient descent procedure  is carried out for an increasing number of iterations $n$.
    
    The hypothesis is that if the gradient descent procedure is capable of approximating $\hat y$, then (\ref{eq10}) holds.
    
    \begin{equation}
        a_{n+1} = a_n - g * \epsilon, \; n \to \infty \implies ||O(a, \theta) - O(a_g, \theta) ||_2 \to 0
        \label{eq10}
    \end{equation}
    
    For a more precise measure of the generated input's closeness to the target input $a$, we defined $m_i$ as the $L^2$ norm of the difference between generated and target inputs in Equation \ref{eq11}, analogously to what was done to measure output tensor differences.  
    
    \begin{equation}
        m_i = ||a - a_g||_2
        \label{eq11}
    \end{equation}

\begin{figure}[h]
        \centering
        \includegraphics[width=0.85\textwidth]{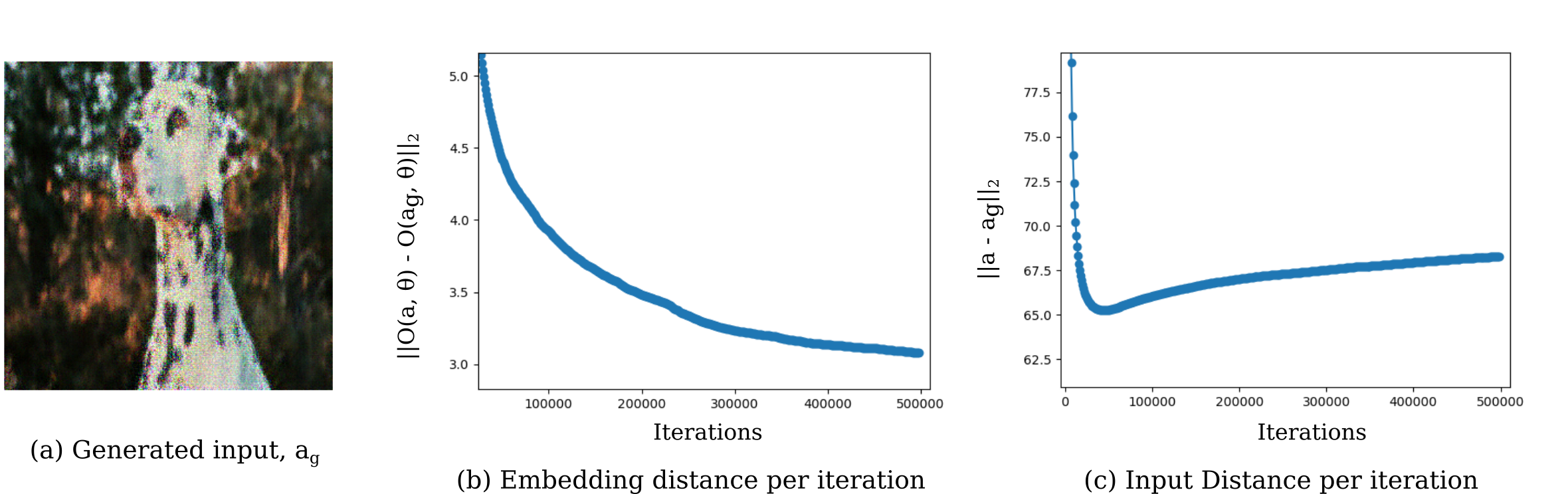}
        \caption{Untrained ResNet50 Layer Conv2 Distances. Experiment conducted with a variable $\epsilon$, scaled linearly from some constant $c$ to 0 as iterations increase.}
        \label{fig4}
    \end{figure}

    We investigate this first in layer Conv2 with an untrained ResNet50, iterating the non-Gaussian convolved version of gradient descent (\ref{eq5}). The results are shown in Figure \ref{fig4}, where it is observed that an exponential decay in the output distance between generated and original embeddings, $||O(a, \theta) - O(a_g, \theta)||_2$, occurs even as there is an increase in input distance $m_i$. 
    
\begin{figure}[h]
        \centering
        \includegraphics[width=0.85\textwidth]{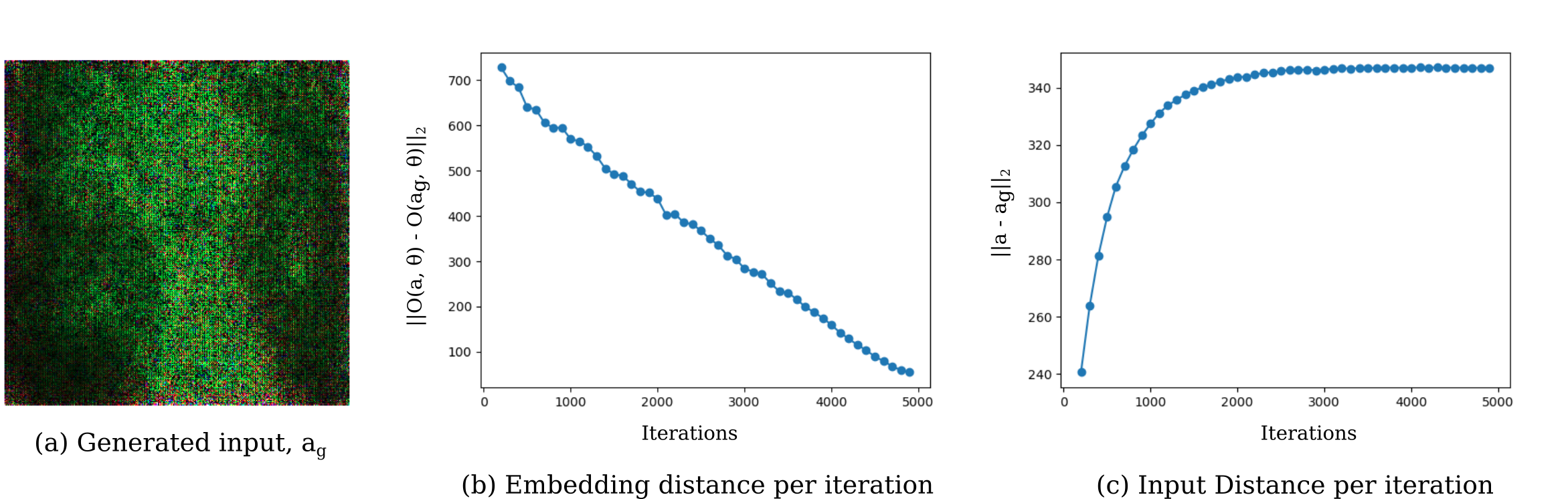}
        \caption{Untrained ResNet50 Layer Conv5 Distances. Experiment conducted with a variable $\epsilon$, scaled linearly as iterations increase from some constant $c$ to 0.}
        \label{fig5}
    \end{figure}

    This decoupling of a decrease in $m_g$ with a decrease in $m_i$ is often more pronounced for deeper layers: for ResNet50 layer Conv5 plotted in Figure \ref{fig5}, there is an inverse correlation between the values of $m_g$ and $m_i$ as $n$ increases. 
      
    The finding that an approximation of an output vector $m_g$ can become better as the input space $m_i$ approximation becomes worse suggests non-uniqueness of the output embedding $O$, as having many possible inputs give approximately the same output would yield an inverse relation between $m_g$ and $m_i$ if $a_g \neq a$. In the next section we explore the theoretical basis behind non-uniqueness and why depth contributes to this phenomenon.
    
    After training has completed, it may be wondered if the phenomenon of non-uniqueness remains. Testing this question for ResNet50 layer Conv5, we find in Figure \ref{figs1} that there is evidence that training does not remove non-uniqueness of layer representations.
    
\section{Representations worsen with depth due to non-invertibility}

    From an informational perspective, it at first seems counter-intuitive that depth would lead to poorer representations even if all layers of the model in question may be considered to be overcomplete being that they have more nodes than the input has elements.  Another possible explanation for poor representation is that the transformation $O(a, \theta)$ may be expressed as a linear operator of poor conditioning, meaning that even though this transformation may be invertible in theory it is not invertible in practice due to large differences between the smallest and largest eigenvalues of that transformation.  Therefore we examine both exact as well as approximate non-invertibility as potential causes of poor hidden layer representation.  It should be noted that we did not observe arithmetic precision to play a significant role in representation quality at least for the experiments documented here, as for example changing from 32-bit to 64-bit floating precision did not affect representations for convolutional or fully connected models.
    
    First we observe that nearly all deep learning vision models studied today contain non-invertible transformations between successive layers. To see why this is the case, take a typical deep learning vision model such as ResNet.  Between layers we have various operations: convolutions, max and average pooling, with the somewhat special operation (to the ResNet model family) of element-wise tensor addition.  In the case of convolution and pooling operations, observe that any layer with fewer elements than the previous layer cannot uniquely define that previous layer.  For the case of a fully connected neural network with no nonlinearity applied, any layer that has fewer elements than the previous layer is equivalent to a transformation by a non-rectangular matrix (followed by addition of a bias term) and non-rectangular matricies are in the general case non-invertible.
    
    Approximate non-invertibility may be restated as the case where an approximation of some embedding tensor (the representation) may not be a good approximation of the input that gave that embedding.  Say that one were to train a fully connected model with no biases or nonlinearities, where model is invertible such that any given output gave a unique input. The model may be represented by a single transformation $ABCDx = Ex$.  Now suppose that the transformation $E$ contracted or expanded the basis vectors of $x$ at widely different scales, which is a case of poor conditioning of the transformation $E$. This means that points initially either close to or far from some target input $a$ are transformed to lie in the same region $E(a + \delta)$, depending on the direction of $a$ to those points.  If this is the case, it may be necessary to approximate $E(a)$ extremely well in order to make $a_g$ approximate $a$. 
    
\subsection{Arbitrarily accurate input representation is achievable for MLPs with each layer equivalent to an overcomplete transformation}
    
    To investigate the effect of invertibility on input representation accuracy, we focused on input representation without Gaussian convolutions (using Equation \ref{eq5}) applied to simple fully connected neural networks, which for simplicity we call `MLP' architectures.  Unless otherwise denoted, the MLPs used in this work are of typical design (all connections between nodes having a weight parameter, and all nodes except the input containing a bias parameter) but notably no nonlinear transformation is applied to each layer output.  The case of models with ReLU nonlinearity applied to hidden layers is examined in the next section.
    
    First we examine the case of input representation accuracy for a simple three-layer MLP where the number of nodes of the first layer varies.  We use a 3x29x29 down-sampled version of the image of the Dalmatian applied as an input elsewhere in the paper in order to prevent model memory blow-up.  Layer weights are initialized to a Kaiming uniform distribution and the model is untrained.
    
\begin{figure}[h]
        \centering
        \includegraphics[width=0.73\textwidth]{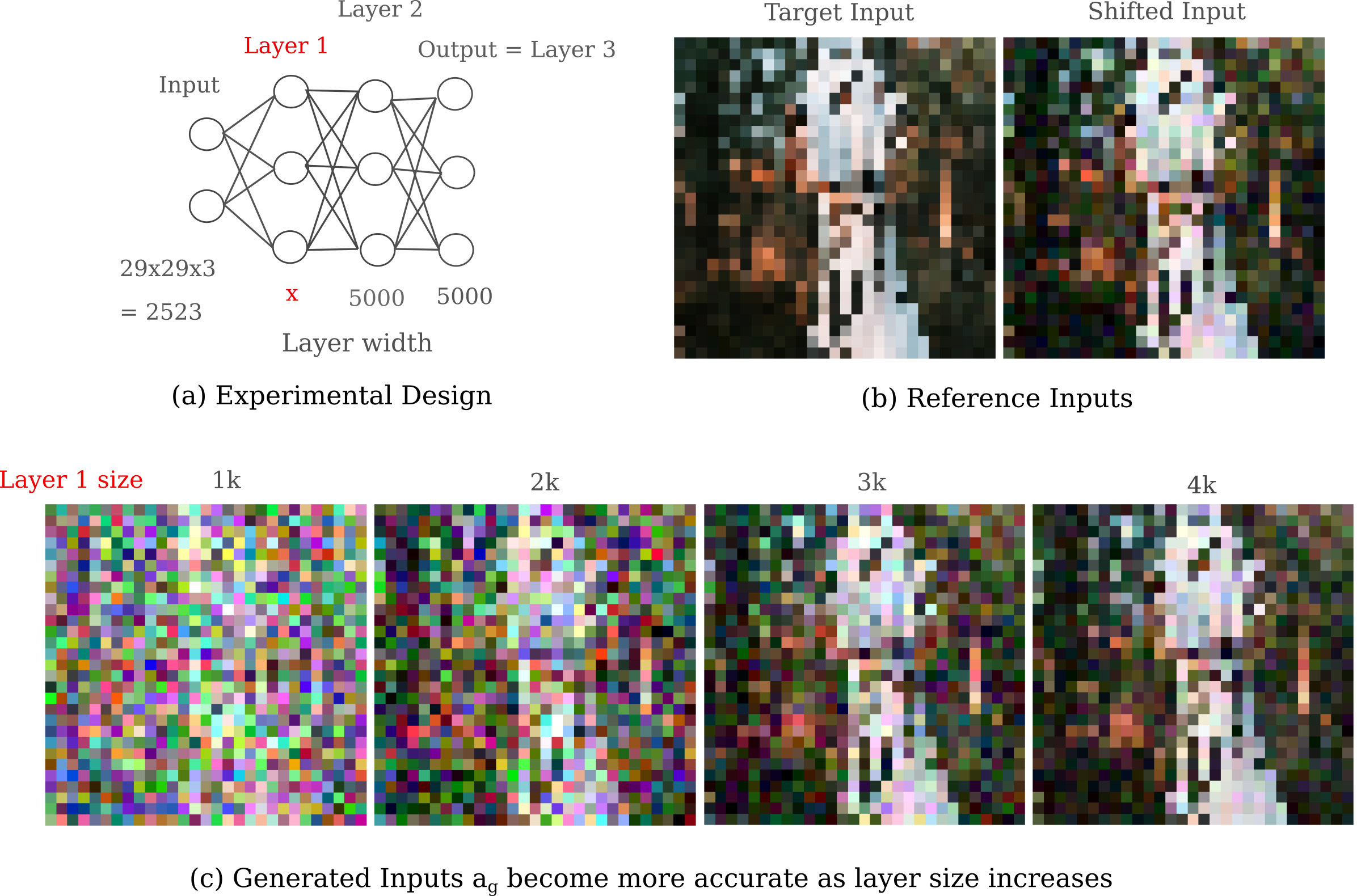}
        \caption{Input representation quality increases with increased first hidden layer width for an untrained MLP.  All generated images yield an output closer to $O(a, \theta)$ than the output of the shifted input $O(a', \theta)$ is.}
        \label{fig6}
    \end{figure}

    Considering the case of exact non-invertibility, the hypothesis is that an arbitrarily `good' input representation is possible if and only if each transformation between successive layers is overcomplete (or in other words if there are at least as many output elements as input elements).  For a fixed number of iterations of (\ref{eq5}), it is evident that the generated input $a_g$ using the output of the final layer as the target embedding becomes more similar to the target input $a$ as the number of elements in the first hidden layer surpasses the number of elements in the input (Figure \ref{fig6}).  
    
\begin{figure}[h]
        \centering
        \includegraphics[width=0.85\textwidth]{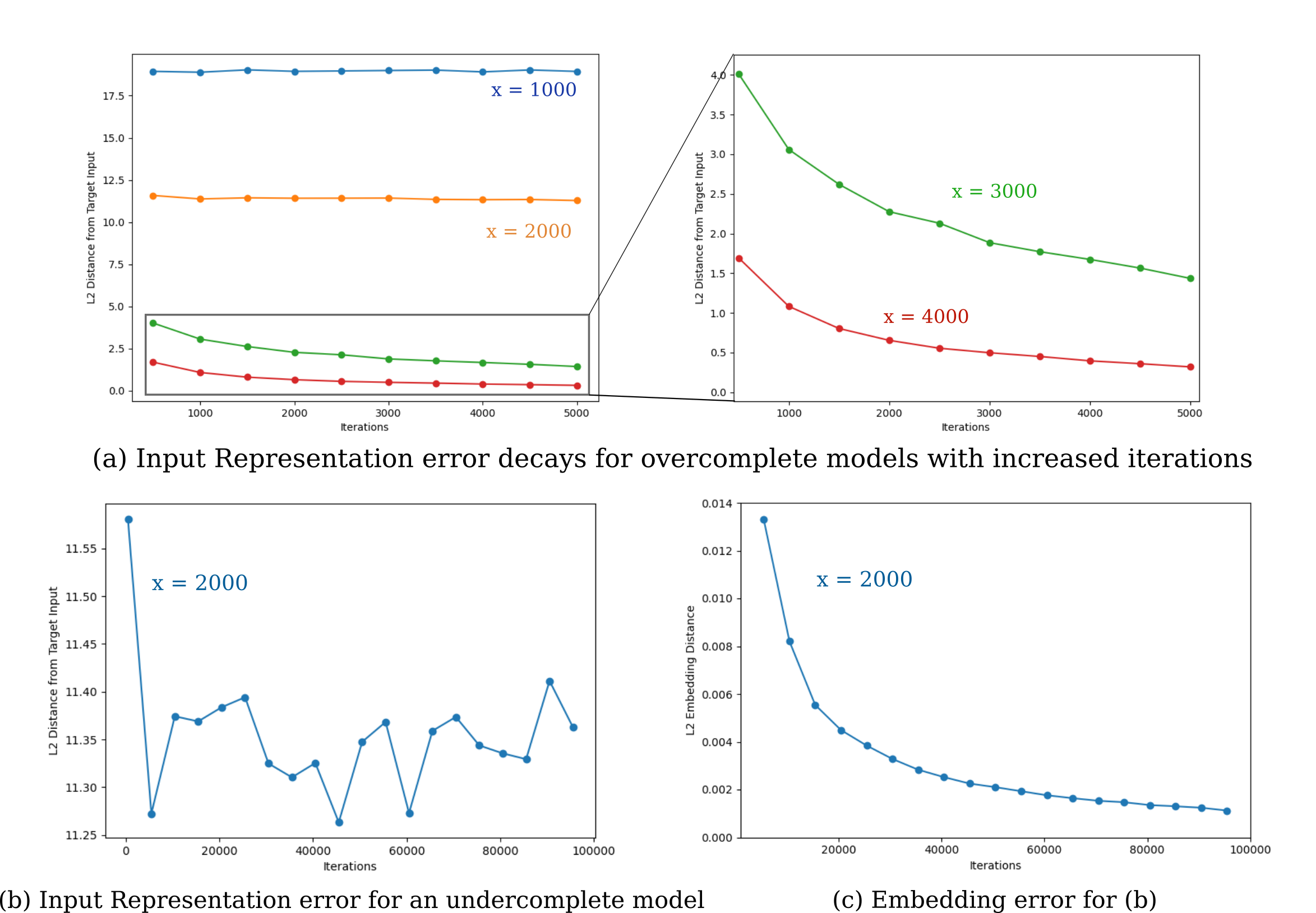}
        \caption{Input representation error heads towards the origin for untrained MLPs with more but not fewer nodes per layer than there are elements in the input.}
        \label{fig7}
    \end{figure}

    The possibility remains that an increase in the number of iterations of (\ref{eq5}) would allow undercomplete models (with layer width $x < 2523$) to approximate $a$ arbitrarily well despite non-invertibility if there were only one input near the starting point $a_0$ that yields an accurate approximation of the desired embedding (even if there are many inputs that equivalently give the same representations, but these may be very far from $a_0$ such that they would not be found via gradient descent). This idea can be easily tested, however, and in Figure \ref{fig7} is observed to not be supported for $x=2000$ and $x=1000$: as seen for ResNets, increasing iterations of gradient descent (\ref{eq5}) yield better and better output approximations, i.e. $||O(a, \theta) - O(a_g, \theta)||_2 \to 0$, but the input representation distance $||a - a_g||$ becomes slightly larger and certainly does not decay towards the origin.
    
    To conclude this section, we obtain experimental evidence for the idea that non-invertibility of a multi-layer perceptron results in necessarily imperfect input representation as a result of non-uniqueness. It should be noted that non-invertibility does not necessarily lead to the poor representation phenomenon observed in the last section, however: take for example the case where of all possible inputs that yield an identical output only one is near $a_0$.  
    
\subsection{Approximate non-uniqueness contributes to poor representation clarity but does not account for asymptotically poor representations}
    
    The inability to accurately represent an input given non-invertible layer transformations does not preclude the idea that deep layers have poor representations of their inputs in part because of approximate non-uniqueness.  To investigate this possibility, an experiment was designed in which an MLP model was constructed to be identical to the case for Figure \ref{fig6} except that the number of layers initialized were changed, rather than the width of the first hidden layer.  
    
    \begin{figure}[h]
        \centering
        \includegraphics[width=0.73\textwidth]{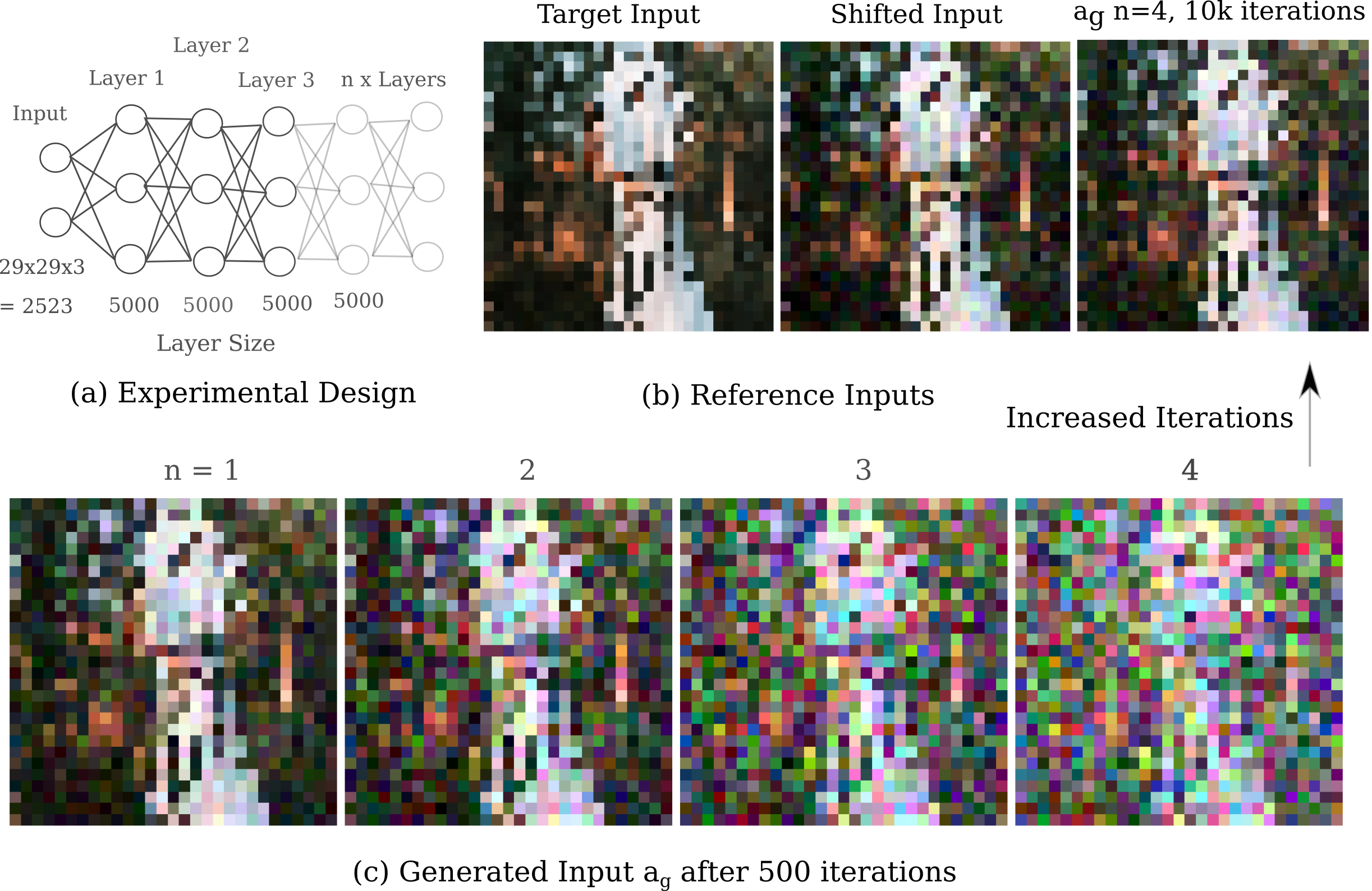}
        \caption{Approximate non-invertibility with increasing depth in an untrained MLP.}
        \label{fig8}
    \end{figure}
    
    As shown in Figure \ref{fig8}, an increase in the number of layers between the input and output leads to less accurate representation visualizations of the output layer for a fixed number of gradient descent iterations.  But if the number of gradient descent iterations grows, deeper models form representations that are capable of approximating the input well (Figure \ref{fig8}).
    
    It may be wondered exactly how poorly conditioned (or equivalently how approximately non-invertible) an invertible fully connected model is.  Rather than investigate this directly by solving for a model's eigenvalues, a perhaps more intuitive way of approaching this problem is to observe what possible changes can be made in the input for some small change in the output.
    
    For simplicity, we consider a model where $A, B, C$ are matricies corresponding to the layer weights, and all layers have a width equal to the number of input elements, $2523$. Given this model, we can invert the last layer as follows: given the output vector $o$ and layer input vector $x$, we can solve for $x$ in terms of $o$ and weight matrix $C$ and bias vector $b$ as given in Equation (\ref{eq13}).
    
    \begin{equation}
    \begin{gathered}
        o = Cx + b \implies \\
        x = C^{-1}(o - b)
        \label{eq13}
        \end{gathered}
    \end{equation}

    Applying (\ref{eq13}) iteratively to layers $C, B, A$ we can recover the input given any output vector.  Given input $a$, it can be experimentally verified that a fully connected three-deep model may be inverted accurately using Torch library matrix inversion, but notably only if double (64-bit floating) precision is used.  If we add a small random normal shift to the output we can then recover the input $a''$, which is defined as $O(a'', \theta) = O(a, \theta) + \mathcal{N}(O(a, \theta); \mu=0, \sigma=1/1000)$.  The distance $||a - a''||$ can then be compared to our standard shifted input distance $||a - a'||$, being that the output distance $||O(a', \theta) - O(a, \theta)) > || O(a'', \theta) - O(a', \theta)||$ by design.

    From Figure \ref{figs2} it can be appreciated that nearly all possible inputs $a''$ are much farther (more than one million times to be specific) from $a$ than $a'$, meaning that even an untrained three-hidden-layer MLP is typically quite poorly conditioned.  
    
\subsection{ReLU nonlinearities make overcomplete architectures require more elements per layer than the previous layer}
    
    All of the MLP experiments in this work so far have employed typical architectural choices (weights and biases) but contain no nonlinear activation function. While experimenting with various MLP architectures, it became clear that including ReLU activations, and to a lesser extent other nonlinear activations, after each layer leads to a substantial drop in representation visualization accuracy, or more precisely $|| a - a_g ||$ increased for a given output distance.  ReLU is given in Equation \ref{eq11} for convenience.
    
    \begin{equation}
       y = f(x) =  
      \begin{cases}
        0, & \text{if } x \leq 0 \\
        x, & \text{if } x > 0 \\
      \end{cases}
      \label{eq11}
    \end{equation}
    
\begin{figure}[h]
        \centering
        \includegraphics[width=0.6\textwidth]{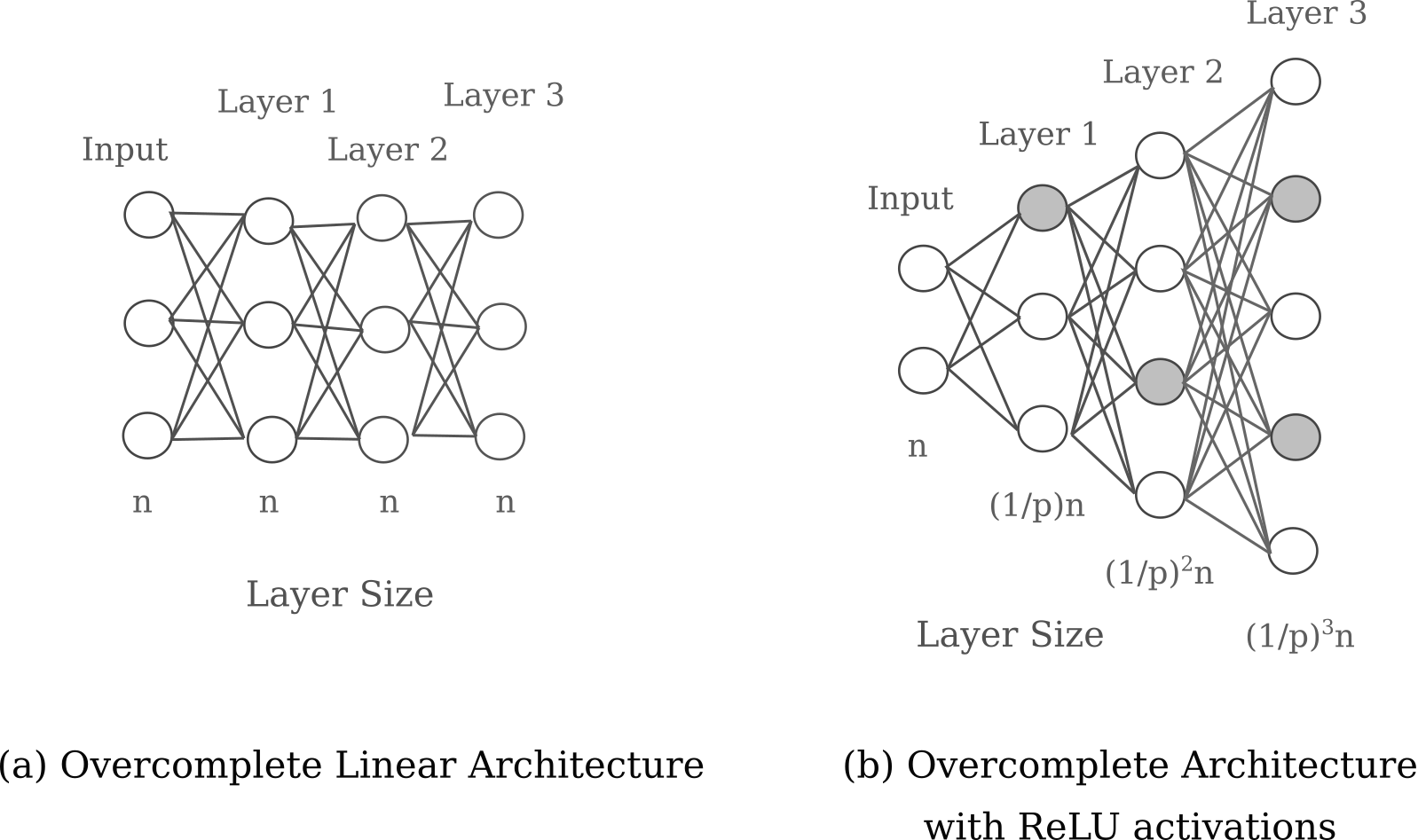}
        \caption{ReLU activations necessitate that each layer be larger than the previous in order to be capable of copying an input.}
        \label{fig9}
    \end{figure}

    Considering the nature of what is required for invertibility between layers, it is apparent that adding ReLU to a layer's output diminishes that layer's ability to represent the layer's input simply because some number of the elements that were possibly unique before ReLU transformation are now equivalently zero, such that no information on the exact inputs of these elements may be obtained.  More precisely, all elements of the set of layer outputs ${y_0, y_1, y_2, ...}$ less than zero are now identically zero, $f({y_i: y_i < 0}) = {0}$.  If we assume that there is some probability $p$ such that $p(y) = P(y: y < 0)$ and assume further that this probability does not change for different layers, it is clear that the number of output elements required in order to solve for the input elements is $1/p$ which is evident from the definition of the expectation of $p(y)$.  Extending this to the case of multiple layers, we find that for a model with ReLU nonlinearities to be overcomplete this model must have an increasing number of neurons per layer, and the precise number of neurons required for layer $n$ is $m(1/p)^n$ where $m$ denotes the size of $a$.  This is summarized in Figure \ref{fig9} for clarity.
    
    At the start of training, it may be assumed that half of each layer's neurons are zeroed out if ReLU is applied.  If this is the case, each successive layer must double in width in order to be capable of preserving all the information necessary to represent the input exactly.  It should be noted that for trained model it is rarely seen that there is some constant probability of ReLU transforming an activation to 0 for different layers, and that a model may be chosen that can approximate perfect representation with only a constant number of neurons per layer regardless of whether or not any nonlinear activation used if all activations are greater than 0. 
    
    This observation provides a theoretical basis for the standard autoencoder architecture of connecting the latent space hidden layer to the output via layers each larger than the last: given that such models are typically trained with ReLU activation applied to hidden layers, increasing the widths of layers connecting the latent space to the output is required in order to prevent information loss.
    
    \subsection{Why representation accuracy decreases with depth}
    
    Returning to ResNet, it appears that the non-uniqueness due to absolute non-invertibility as well as approximate non-invertibility both contribute to poor representation accuracy in deep layers.  Evidence for absolute non-invertibility is the observation that the limit of the distance between the approximation of $O(a, \theta)$ and the target output itself tends towards the origin even as the distance $||a - a_g||_2$ increases.  Poor conditioning would make $O(a, \theta)$ difficult to approximate but would not expected to lead to consistently larger $||a - a_g||_2$ distances upon better approximations of the output, whereas exact non-uniqueness certainly would.  On the other hand, it typically takes a very large number of iterations at very small learning rates for this distance to approach the origin, which is what would be expected given approximate non-invertibility.
    
    A lack of invertibility between layers means that a potentially infinite number of linear combinations of the input can give one output.  As realistic image inputs are bounded in some way (for the inputs used in this work, every input element is between 0 and 1, ie $a_i \in [0, 1]$), we can expect a finite number of inputs to yield some output when given to the model. Say for example that a model has 5 layers and each layer has 100 possible inputs for a given output.  For example, if 100 distinct inputs for layer 1 yield a given output, the output of the second layer there are $100^2$ possibilities and for layer $n$ we have in general $100^n$ possibilities. 
    
    Furthermore, exact representation is further hampered if a model contains distributed representations because subsets of the input may be replaced with linear combinations of subsets of the input without changing the output.
    
    In conclusion, most deep learning architectures applied to image recognition can be shown to exhibit non-unique representations.  This non-uniqueness results in representations that lose input information such that they are unable to exactly copy the input.  Once non-uniqueness is removed, arbitrary approximation of an embedding yields arbitrary approximation of the target input.  For representations with a fixed number of iterations, however, approximate non-uniqueness does limit representation accuracy.
    
\section{Hidden layers transform arbitrary inputs into expected ones.}

\begin{figure}[h]
        \centering
        \includegraphics[width=0.99\textwidth]{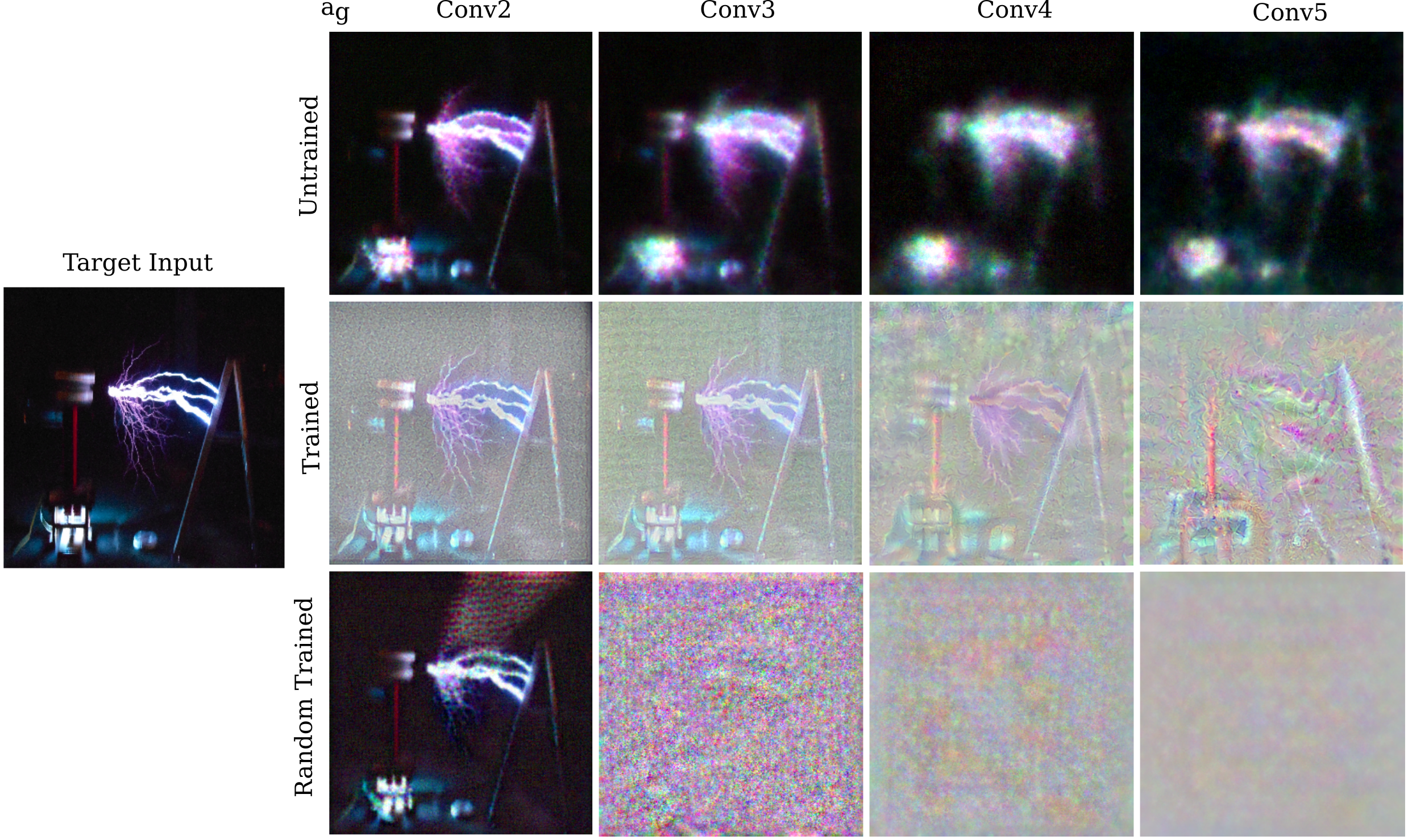}
        \caption{Representation visualizations of an object (Tesla Coil) not present in the ImageNet dataset using ResNet50. In (c) note how a model that has been trained to identify labelled random Gaussian noise maps the input to that noise in the middle and deeper layer representations.}
        \label{fig10}
    \end{figure}

\begin{figure}[h]
        \centering
        \includegraphics[width=0.95\textwidth]{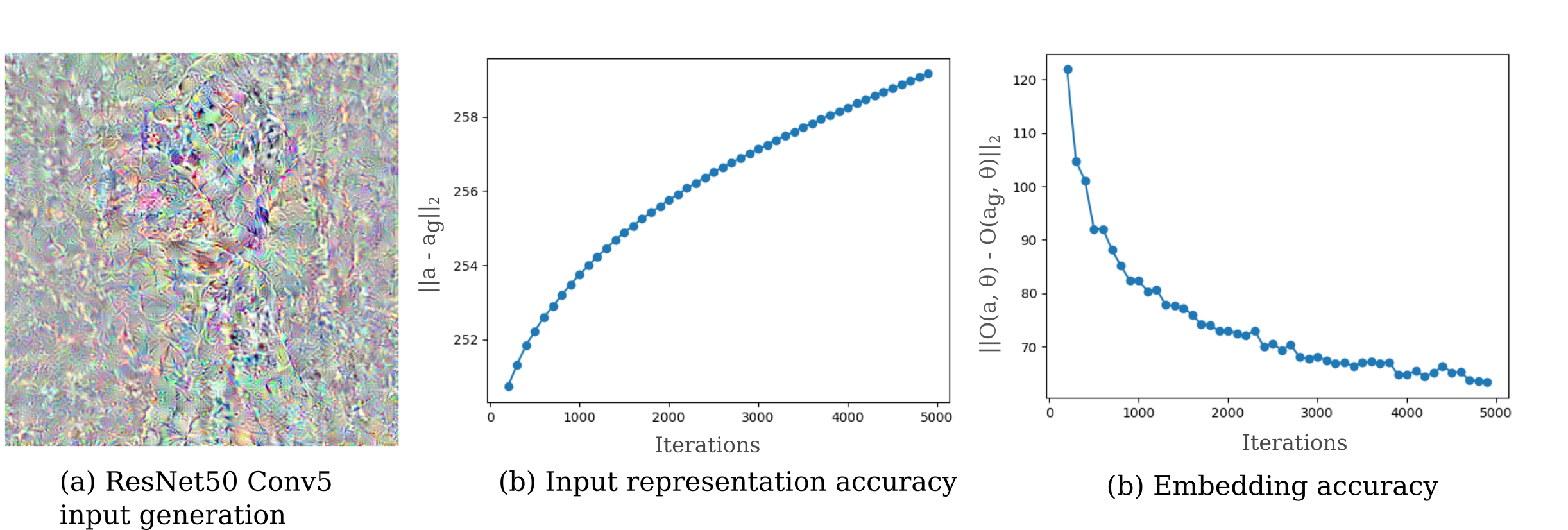}
        \caption{Trained ResNet50 input representations are non-unique.}
        \label{figs1}
    \end{figure}

\begin{figure}[h]
        \centering
        \includegraphics[width=0.9\textwidth]{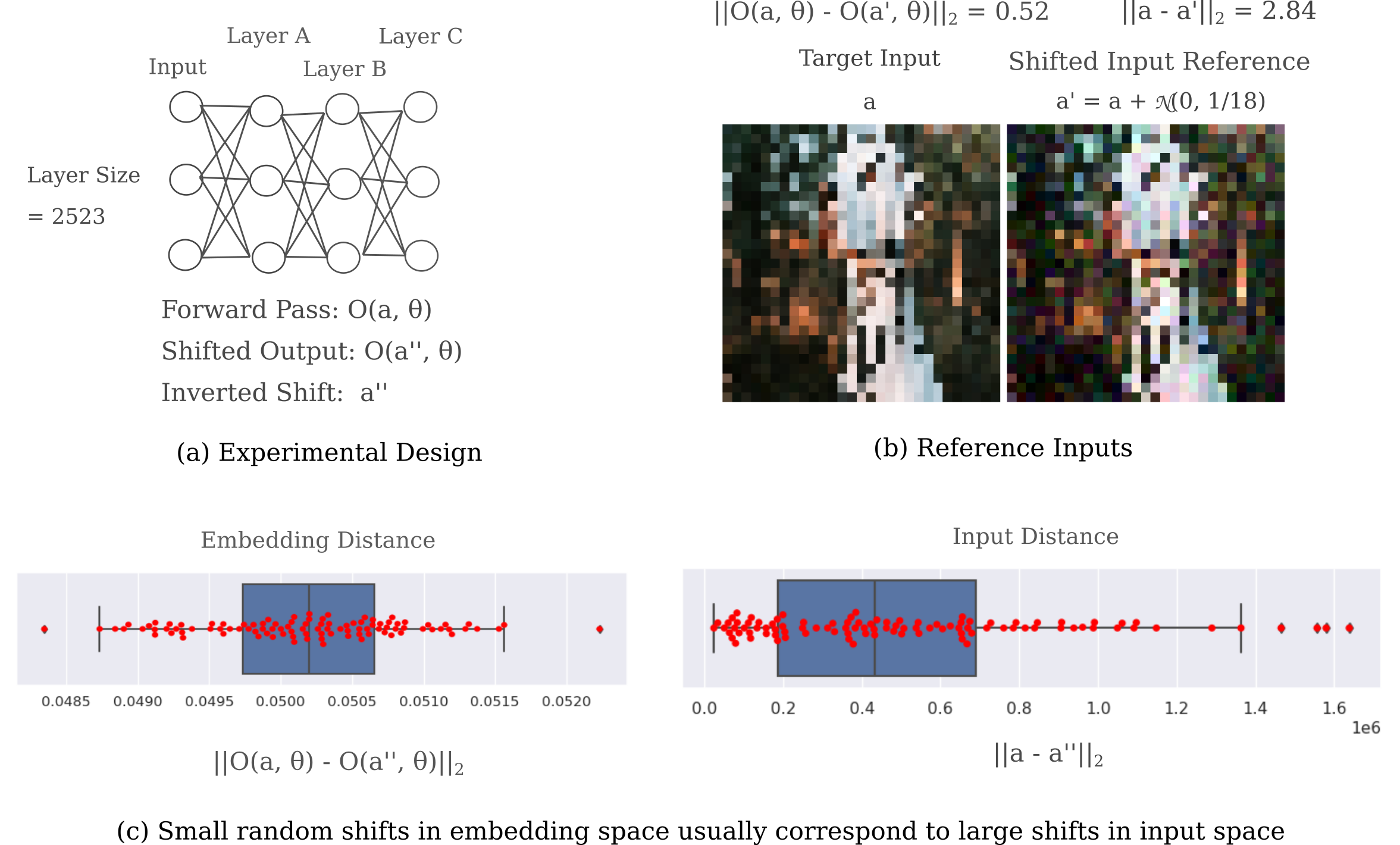}
        \caption{Inverting a slightly shifted output yields a very large input shift for an untrained MLP.}
        \label{figs2}
    \end{figure}

\begin{figure}[h]
        \centering
        \includegraphics[width=0.8\textwidth]{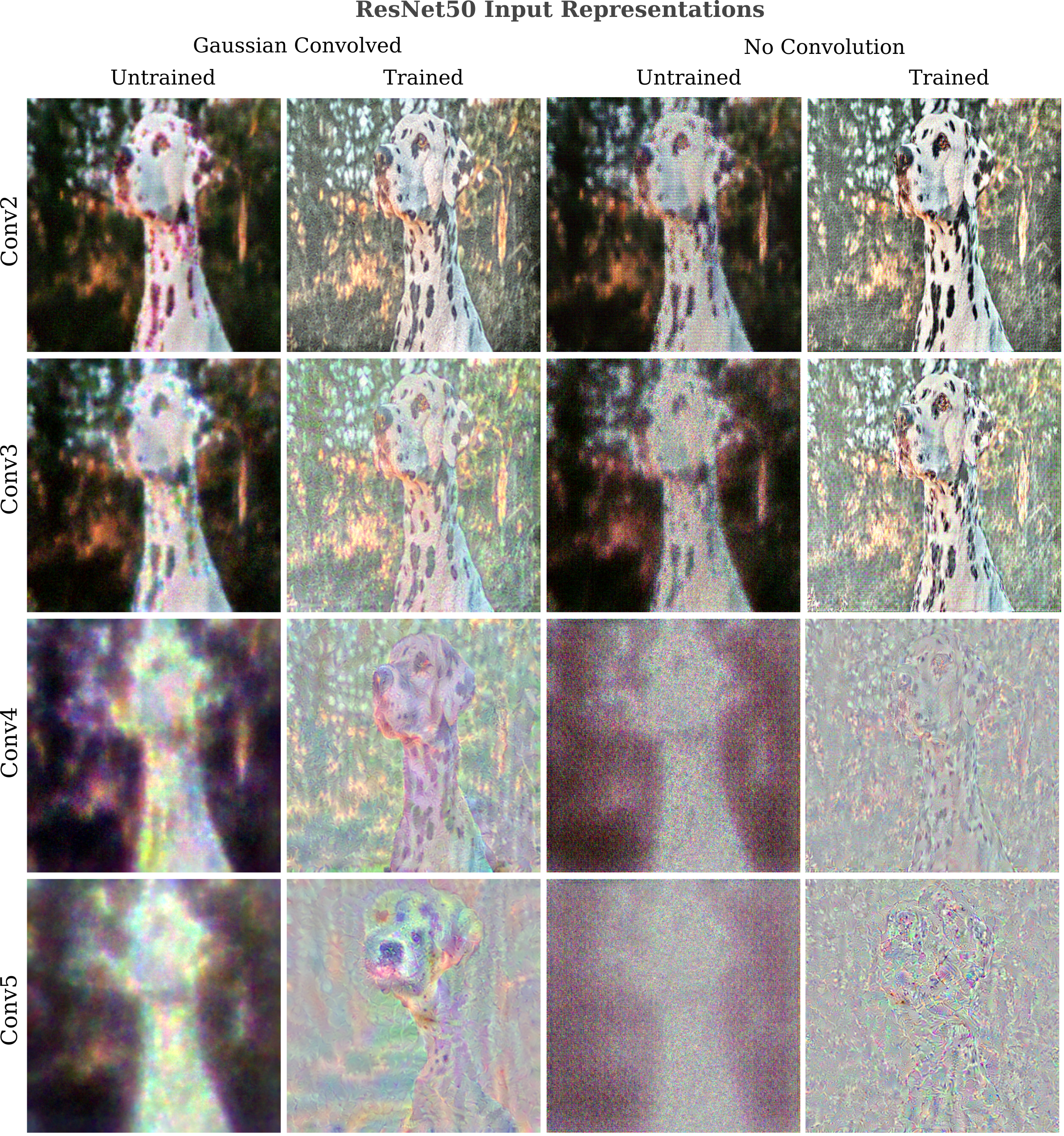}
        \caption{ResNet50 Input Representations before and after training on ImageNet with or without Gaussian convolution in the generation process.}
        \label{figs3}
    \end{figure}

    Comparing Figures \ref{fig1} and \ref{fig2}, it is evident that training leads to an increase in representation clarity in early layers for a target input that is a member of a class in the training dataset, regardless of whether Gaussian convolution was used in the representation visualization process (Figure \ref{figs3}).  It may be wondered how invariant this increase in clarity is: does it also apply to images that are not members of the training dataset in question, and does any training protocol yield greater clarity in early layer representations?  The first question is addressed using an image of a Tesla coil, which is not a member of the ImageNet dataset and would not be observed during training. We see in Figure \ref{fig10} that indeed there is still a marked increase in clarity in the ImageNet-trained model's representation of this input compared to the untrained model.
    
    It may be hypothesized that recognizing any image would lead to a clearer representation after training, but we find that this is not the case.  Training ResNet50 to classify 10k images of random Gaussian noise $\mathcal{N}(a; \; \mu=1/2, \; \sigma=1/3)$ randomly labeled as one of 100 classes, we find that layers Conv3 and Conv4 are noticeably less clear than for the same layers either untrained or trained on ImageNet (Figure \ref{fig10}).  Comparing the kernal weights learned by training on ImageNet versus training on random noise, it is observed that the latter do not contain the the wavelet transformations that are typical of the former (Figure \ref{figs5}).
    
\begin{figure}[h]
        \centering
        \includegraphics[width=0.85\textwidth]{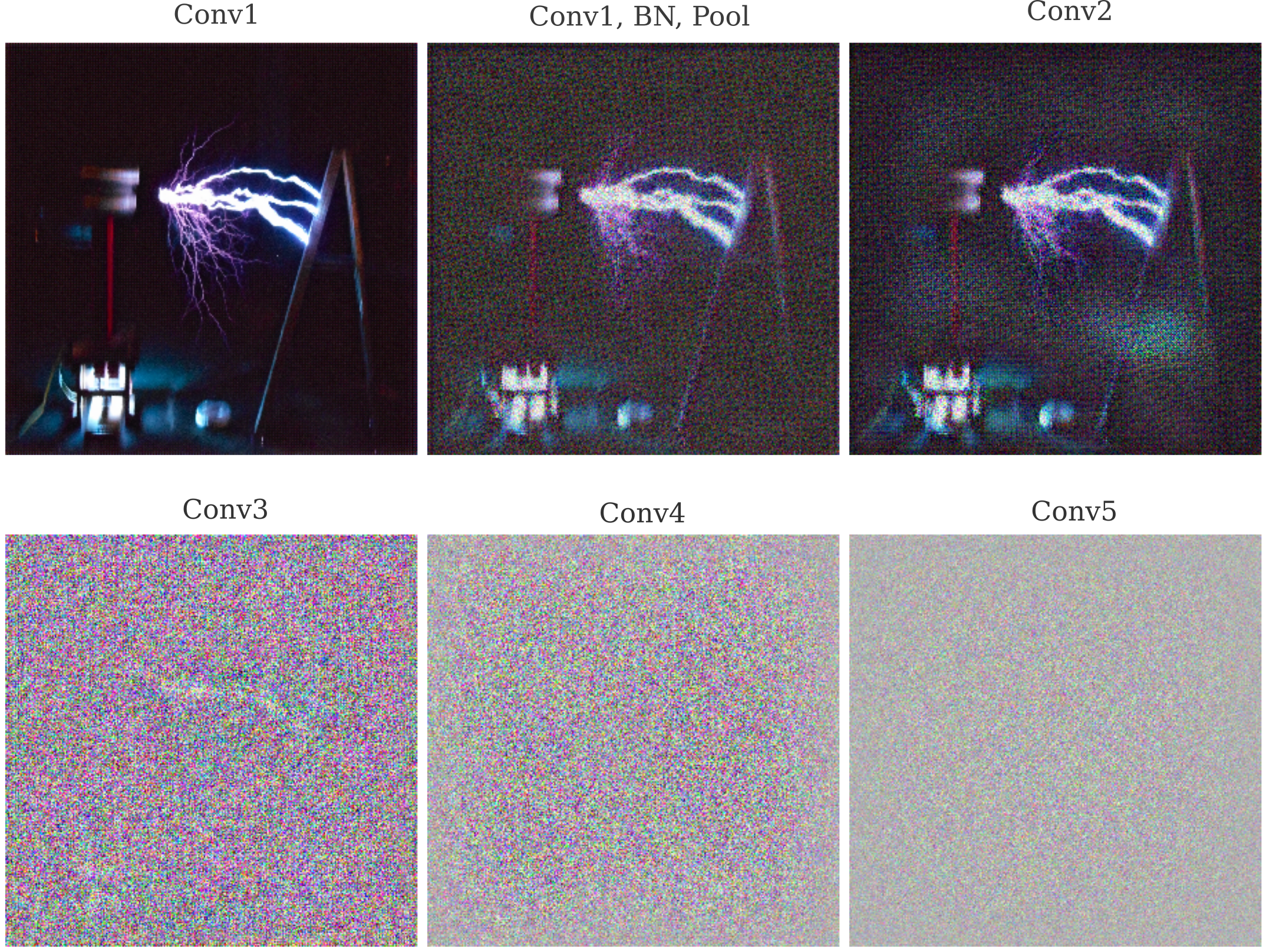}
        \caption{Tesla coil input representation of ResNet50 trained on fixed random samples, without Gaussian convolution.  All representations have a closer embedding distance than the standard shifted input $a' = a + \mathcal{N}(a; \mu=0.7, \sigma=1/18)$}
        \label{figs4}
    \end{figure}

    It is interesting to note that the input representation visualizations for this random-trained model \ref{fig10} do not give very accurate outputs, or more specifically $||O(a_g, \theta) - O(a, \theta)||$ does not become arbitrarily small as the number of iterations increases.  This is not the case if Gaussian convolution is omitted, although the representation visualizations are qualitatively very similar (Figure \ref{figs4}).  When one considers that the training data for the model in question had no inputs that are well-fit by a convolved version of themselves, this finding is not surprising.
    
    Thus a model trained to recognize images of random noise (and recognize them successfully) does not noticeably sharpen the input representation in early layers, and actively transforms natural image representations into those nearly indistinguishable from the pattern of noise present in the training set.  This provides evidence for the notion that rather than simply selecting for certain features in an input, deep learning hidden layers map inputs to representations of inputs that the model has learned to expect from the training data. If deep learning models are capable of approximating the low-dimensional manifold defined by the set of inputs \citep{cayton2005algorithms}, deep hidden layers are here observed to map arbitrary inputs to the manifold learned even if the actual input is very different from that manifold. This idea is a modification of the standard manifold learning hypothesis in which inputs are assumed to exist on a low-dimensional manifold embedded in high-dimensional space such that the model learns to perform mappings on this manifold.  Instead, we find that arbitrary inputs (which are not necessarily on the manifold) are mapped to the manifold or manifolds learned by hidden layers, regardless of whether the inputs are actually on the manifold or far from it.

\section{Implications for representation learning}

\begin{figure}[h]
        \centering
        \includegraphics[width=0.85\textwidth]{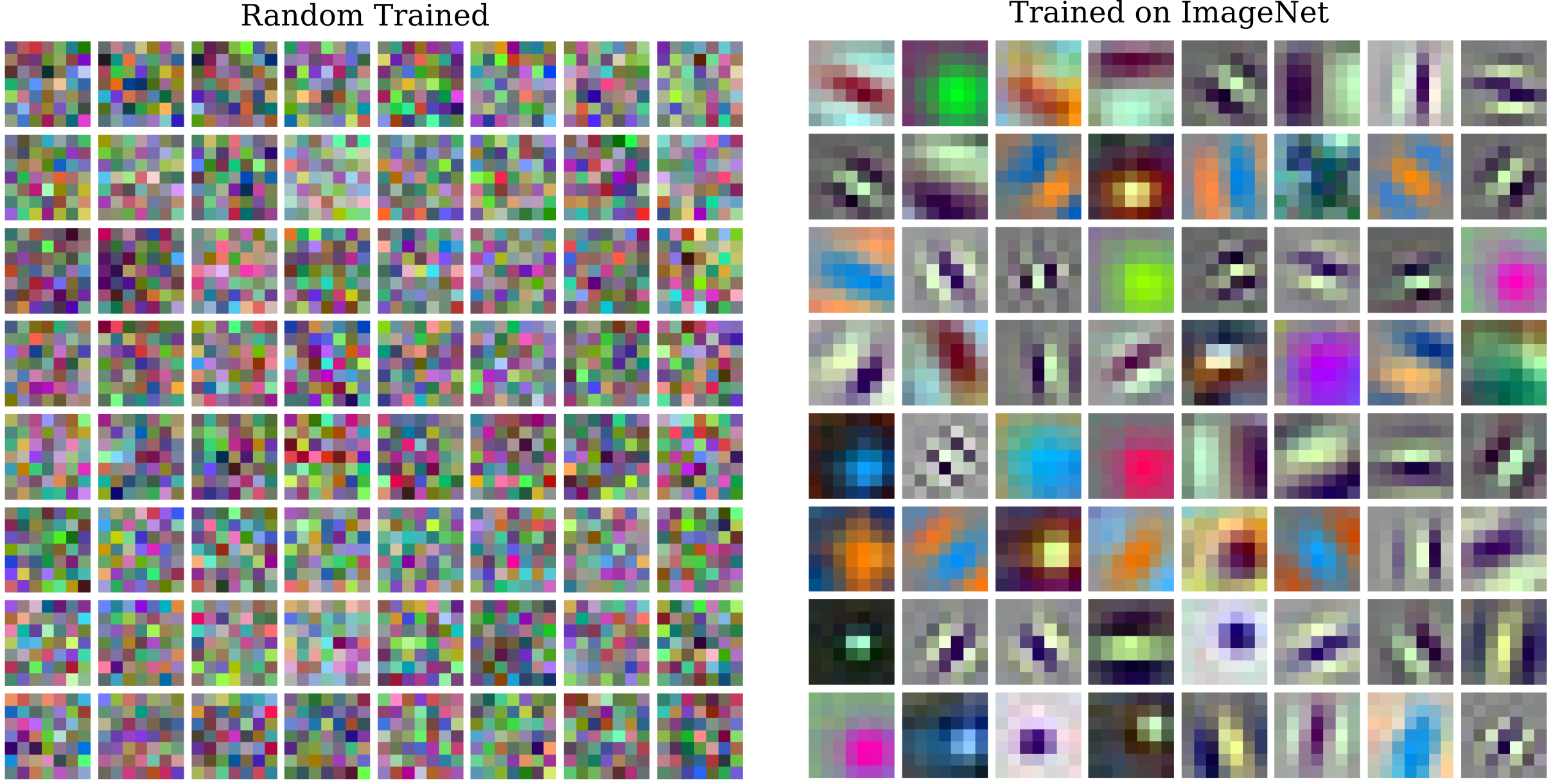}
        \caption{ResNet50 Conv1 convolutional kernal weights form Gabor function wavelets after training on natural images but not random Gaussian images.}
        \label{figs5}
    \end{figure}

    The convolutional kernals that are learned by early layers during image recognition training resemble kernals hand-designed for edge detection, and it can be appreciated kernals of ResNet layers Conv1 (after max pooling), Conv2, and Conv3 appear to be effective at sharpening the representation relative to what it was at the start of training. It has been appreciated for some time that convolutional models often learn kernals in their early layers that resemble the Gabor functions that are the products of trigonometric wavelets and Gaussian functions \citep{mehrotra1992gabor, goodfellow2016deep}. Indeed, ResNet50's layer Conv1 kernal weights are observed to learn Gabor and other wavelet functions after training on ImageNet (Figure \ref{figs5}).  Wavelets provide a basis for locally efficient encodings of sharp edges \citep{gowers2008princeton} and as such may be capable of preventing some of the blurring that is observed to occur for non-unique transformations typical of pooling or strided convolutional layers. We postulate that wavelet functions are learned precisely because they create an output that is more unique with respect to certain aspects of the input than the naive output, resulting in representations from these layers to become visually sharper after training on ImageNet.  
    
    We also offer an explanation for the recently observed difference between early and late layer convolutional filter correlation \citep{Raghu2021}.  Early layers attempt to restrict the possible feasible (ie $a_{m, n} \in [0, 1]$) inputs that yield some given embedding, but late layers do not and instead attempt to map the early layer representations to the manifold learned during training.

    It has been observed that image recognition models such as ResNets are capable of memorizing pure noise \citep{zhang2021understanding}, which is somewhat counter-intuitive given their resistance to overfitting natural images.  In light of the results provided in this work, it seems likely that the process of learning to recognize pure noise is very different than the process of learning to recognize natural images given the difference in ability to represent the input after training, even if both lead to training accuracy at unity.  Other work has also suggested a similar distinction \citep{anagostidis2022}.
    
    Finally, our work provides an explanation for the often-observed requirement for specific types of regularizations such as Gaussian convolution and position invariance when an output class is used to represent an input. It is clear that many possible images may represent an output (to any degree of accuracy specified) due to non-uniqueness, and there is no guarantee that these images resemble natural ones that are examples of the class in question. Mandating the generated images to have some statistical similarity to natural images is required to prevent non-uniqueness from leading to unrecognizable image generation.
    
\section{Conclusion and Future Perspectives}
    In this work we have explored the ability of various layers of deep learning vision models to represent some input. Information on the input is lost during the forward pass, replaced in later layers with the model's expectation of what the input resembles from information gained during training. 
    
    Inexact representation might be expected to be one reason why deep learning models generalize so well even when they have far more nodes than are required for memorization.  If this were true, one would expect for an `inverted' architecture that is capable of exact image representation (for example, a fully connected architecture with an increasing number of nodes per layer) to be more prone to overfitting relative to the standard architectures used for image classification where the number of nodes decreases with increasing depth.  
    
    But this is not observed to be the case: for both CIFAR10 and CIFAR100, MLP-stype architectures that with inverted architectures overfit no more readily than standard architectures.  This is also true for convolutional models applied to the same datasets, suggesting that some other regularizer may be present to prevent overfitting even for a model that is capable of precise input representation in all hidden layers.  But as the gradient flows from the loss of the last layer to the criterion, it is also possible that precise input representation in late hidden layers is not relevant for classification model overfitting given that the output will always be limited in how it can represent the input, assuming fewer classes than input elements.
    
    The results presented in this work appear to be general to other convolutional image classification models (GoogleNet etc.) but it remains to be seen whether they are also applicable to vision transformers, mixers, or other non-convolutional models.

\bibliographystyle{unsrtnat}
\bibliography{references}  

\beginsupplement
\section{Appendix}

    Experiments for this work were performed in Colab. Code and target images used in this paper may be found in \url{https://github.com/blbadger/depth-representation}.  Models were modified from PyTorch implementations, with default ImageNet 1k -trained weights from the Torchvision hub used for the 'trained' models.  'Untrained' models were Kaiming uniform-initialized by default (as they were constructed with PyTorch).
    
    The representation visualization procedures (\ref{eq5}) and especially (\ref{eq6}) were found to be sensitive to changes in the learning rate employed. Typically shallow layer visualization were obtained with larger learning rates than deep layer visualization, with $\epsilon$ ranging from approximately $1/2$ for Conv1 to $1/1000$ for layer Conv5.  Learning rates used for the representation visualizations $a_g$ in this work were obtained by random line search, selecting for $\epsilon$ that approximately minimized $||O(a, \theta) - O(a_g, \theta)||$.
    
    For analytic studies on representation visualization with (\ref{eq5}), we note that the value of $\epsilon$ chosen has less of an effect: assuming that it is small enough, there was no observed change in the behavior (ie exponential decay) of the representation accuracy upon increasing or decreasing the learning rate.
    
\end{document}